\documentclass[10pt,twocolumn,letterpaper]{article}

\usepackage{iccv}              
\usepackage[accsupp]{axessibility}  
\usepackage{placeins}
\usepackage{multirow}

\definecolor{iccvblue}{rgb}{0.21,0.49,0.74}
\usepackage[pagebackref,breaklinks,colorlinks,allcolors=iccvblue]{hyperref}

\def\paperID{11} 
\def\confName{ICCV}
\def\confYear{2025}

\title{MAP: End-to-End Autonomous Driving with Map-Assisted Planning}

\author{Huilin Yin$^1$, Yiming Kan$^1$\thanks{This work was supported by the National Natural Science Foundation of China under Grant No. 62433014. Huilin Yin and Yiming Kan are with the College of Electronic and Information Engineering, Tongji University, Shanghai 201804, China (e-mail: \{yinhuilin, yimingkan\}@tongji.edu.cn).
Corresponding author: Yiming Kan (yimingkan@tongji.edu.cn).}, 
Daniel Watzenig$^2$%
\thanks{
Daniel Watzenig is with the Graz University of Technology and the Virtual Vehicle Research, Graz 8010, Austria (e-mail: daniel.watzenig@tugraz.at).} \\
$^1$Tongji University \quad $^2$Graz University of Technology}

\begin{document}
\maketitle
\begin{abstract}
In recent years, end-to-end autonomous driving has attracted increasing attention for its ability to jointly model perception, prediction, and planning within a unified framework. However, most existing approaches underutilize the online mapping module, leaving its potential to enhance trajectory planning largely untapped. This paper proposes \textbf{MAP (Map-Assisted Planning)}, a novel map-assisted end-to-end trajectory planning framework. MAP explicitly integrates segmentation-based map features and the current ego status through a \textbf{Plan-enhancing Online Mapping} module, an \textbf{Ego-status-guided Planning} module, and a \textbf{Weight Adapter} based on current ego status. Experiments conducted on the DAIR-V2X-seq-SPD dataset demonstrate that the proposed method achieves a 16.6\% reduction in L2 displacement error, a 56.2\% reduction in off-road rate, and a 44.5\% improvement in overall score compared to the UniV2X baseline, even without post-processing. Furthermore, it achieves top ranking in Track 2 of the End-to-End Autonomous Driving through V2X Cooperation Challenge of MEIS Workshop @CVPR2025, outperforming the second-best model by 39.5\% in terms of overall score. These results highlight the effectiveness of explicitly leveraging semantic map features in planning and suggest new directions for improving structure design in end-to-end autonomous driving systems. Our code is available at \url{https://gitee.com/kymkym/map.git}.

\end{abstract}

\section{Introduction}

\begin{figure}[t]
  \centering
  \includegraphics[width=1.0\linewidth]{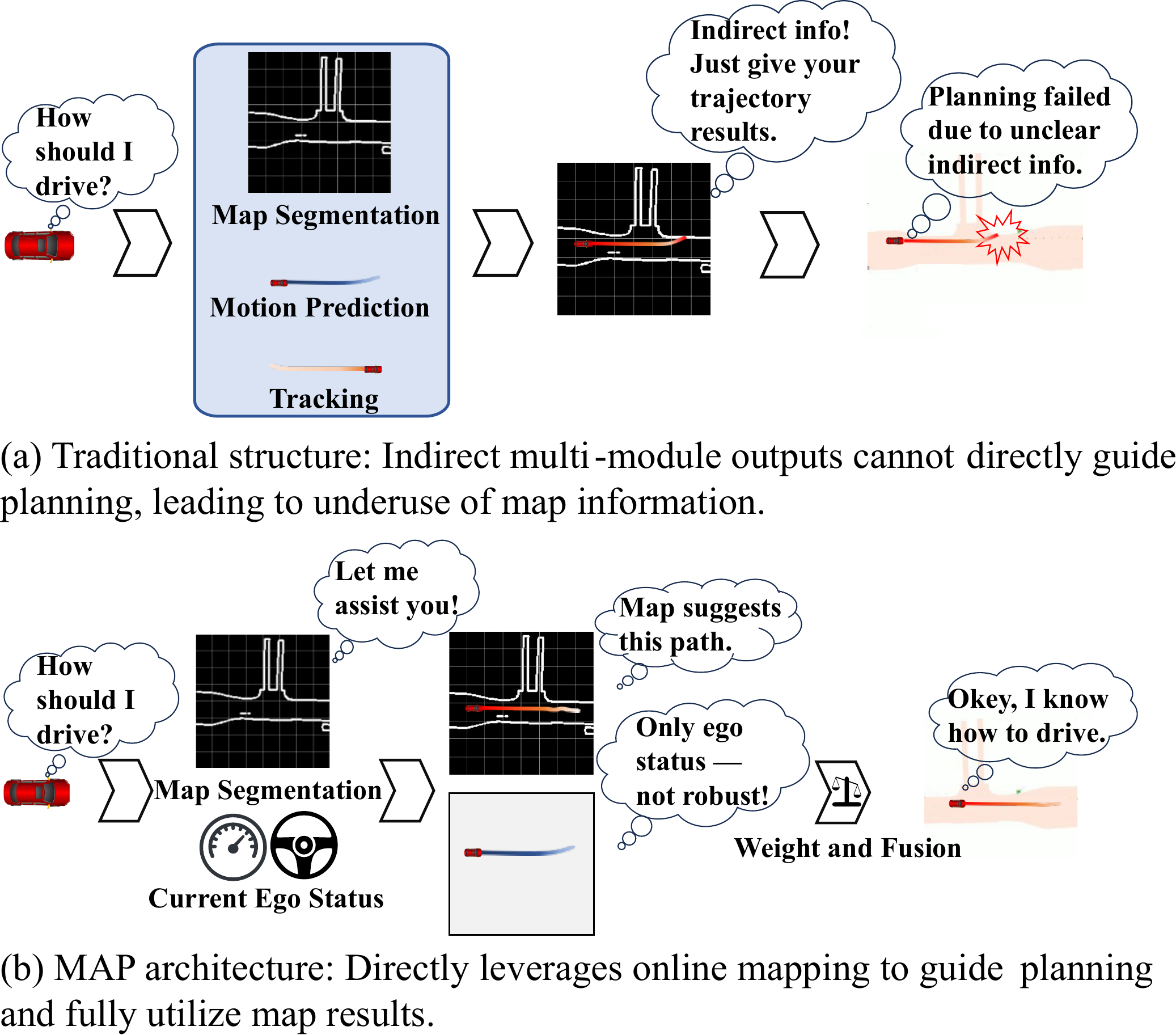}
  \caption{Comparison between traditional and our MAP architecture for end-to-end autonomous driving.}
  \label{fig:overview_fig1}
\end{figure}

Traditional end-to-end autonomous driving systems typically adopt a serial or parallel modular design, where the final trajectory is decoded by a planning module. In such pipelines, the online mapping module is only supervised by task-specific losses, and as more downstream modules are stacked, the contribution of online mapping becomes increasingly vague, as shown in Fig.~\ref{fig:overview_fig1}(a). Whether the segmentation results are accurate or even effectively utilized by the planner remains an often-overlooked question.


From a theoretical perspective, map information provides crucial spatial context for driving decisions, encoding static environmental constraints such as road topology, drivable areas, lane connectivity, and obstacle boundaries. These elements form the foundation of feasible and safe trajectories. Without proper integration of such spatial priors, planning models may rely solely on local dynamics or heuristic rules, leading to brittle or short-sighted behavior. Prior research in classical planning pipelines has long established the necessity of accurate semantic maps for globally consistent trajectory generation, and this remains equally true in end-to-end paradigms. Thus, the absence of a principled mechanism to inject structured map information into planning modules significantly limits the reliability of autonomous driving systems. Our work is motivated by this theoretical gap and aims to systematically enhance planning robustness through explicit map-feature integration.

However, most existing models simply stack modules within a unified architecture, with limited discussion about whether the online mapping module is essential and how its predictions can be effectively exploited. Moreover, although recent studies~\cite{AD-MLP} have shown that comparable performance can be achieved using only historical ego status, further analysis~\cite{Li2024ego} reveals that such approaches are highly sensitive to variations in input ego status, leading to significant degradation in planning performance when the ego status deviates slightly. This highlights the importance of incorporating robust perception cues to ensure safe and reliable driving behavior. We argue that current end-to-end models fall short not because it is unnecessary, but because its outputs are not fully or correctly leveraged in trajectory planning. In this work, we take map segmentation as the entry point and propose to incorporate both semantic map information from the online mapping module and the current ego status to achieve a trajectory planner that is both robust and accurate, as shown in Fig.~\ref{fig:overview_fig1}(b).

To fully realize the potential of the end-to-end paradigm and exploit map information for driving safety, while ensuring that current ego vehicle status contributes to basic planning robustness, we revisit the design of current frameworks and identify three underexplored issues:  
(1) Online mapping is often underutilized, and its intermediate outputs in end-to-end models are rarely checked for correctness.  
(2) The influence of the online mapping module on the final trajectory is not explicitly established, as existing models offer only vague claims about its constraint effects without tightly coupling it with the planning process.  
(3) Most models use segmentation outputs only in post-processing, treating the map module merely as an auxiliary supervision component during training.

To address these gaps, we propose \textbf{MAP (Map Assisted Planning)}, a novel framework that tightly couples map information with trajectory planning. The key idea behind MAP is to leverage complementary strengths of map-based context and ego-vehicle status to improve planning accuracy and robustness. MAP consists of three components: (i) a Plan-enhancing Online Mapping (POM) module that produces trajectory-decodable queries, (ii) a lightweight Ego-status-guided Planning (EP) module guided by the current ego status, and (iii) a weight adapter that adaptively fuses both queries from POM and EP modules.

Specifically, the map-guided planning query is obtained by applying cross-attention between the segmentation features and the current ego status. The ego-guided planning query is generated by cross-attending the current ego status with BEV features. These two queries are then fused using a learned weight from the weight adapter, which encodes the current ego vehicle status via an MLP and outputs a scalar in $[0, 1]$. The final fused query is a weighted combination of the two sources. Furthermore, to promote harmonious cooperation between the two modules and avoid conflicting outputs, only the final decoded trajectory is supervised during training. The intermediate outputs from the EP module remain unsupervised, allowing the system to self-organize the fusion strategy effectively.

Our main contributions are summarized as follows:
\begin{itemize}
  \item We propose a novel end-to-end trajectory planning paradigm, \textbf{MAP (Map-Assisted Planning)}, which explicitly leverages online mapping outputs to assist final trajectory decoding.
  \item We revisit the role of online mapping in end-to-end driving and explore how to combine it with current ego vehicle status for more robust planning through dynamic fusion.
  \item Our method ranks first in Track 2 of the End-to-End Autonomous Driving through V2X Cooperation Challenge at the MEIS Workshop @CVPR2025, achieving state-of-the-art performance in L2 distance, off-road rate, and overall score.
\end{itemize}

\section{Related Work}


\subsection{Online Mapping}
Semantic parsing of HD maps plays a critical role in localization, navigation, and planning. Traditional methods often rely on semantic or instance segmentation in the image domain. Classical architectures like PSPNet~\cite{PSPNet} and the DeepLab series\cite{deeplabv1, deeplabv2v3, deeplabv3plus} have been widely used to detect fundamental map components such as roads, lane boundaries, and crosswalks. More recently, HDMapNet~\cite{HDMapNet} integrates images, LiDAR, and trajectory inputs to produce high-precision BEV semantic maps. VectorMapNet~\cite{liu2022vectormapnet} generates vectorized maps suitable for motion forecasting via clustering and vectorization. MapTR~\cite{Liao2023MapTR} adopts a Transformer-based framework for end-to-end vectorized map construction, enabling direct generation of structured elements like road boundaries and lane lines. As task requirements evolve, Panoptic SegFormer~\cite{panopticsegformer2022} proposes a unified panoptic segmentation framework that fuses semantic and instance segmentation, achieving a favorable balance between accuracy and efficiency. Our work continues to use Panoptic SegFormer on top of the UniV2X baseline for online mapping, while structurally adapting the model for the trajectory decoding task to explore deep integration between map information and end-to-end planning.
\begin{figure*}[t]
\centering
\includegraphics[width=\linewidth]{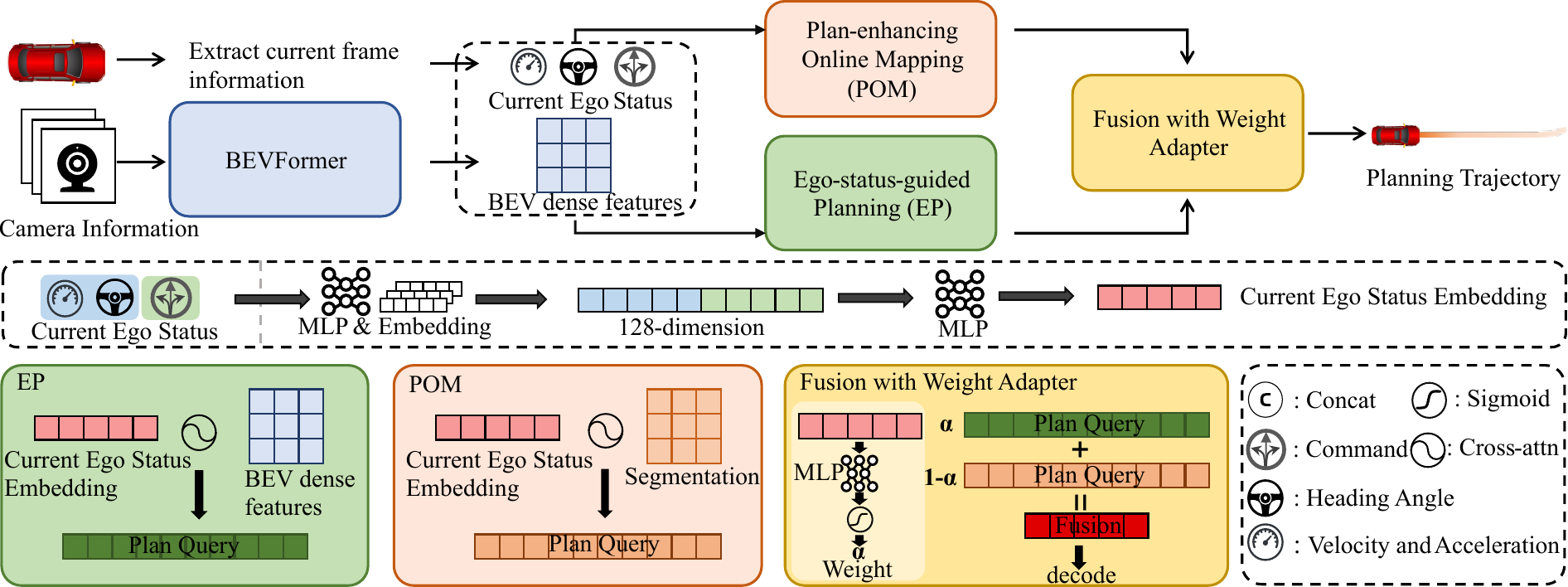}
\caption{Overall model architecture. In the POM module, the map-guided query is obtained via cross-attention between the ego status embedding and the segmentation memory. In the EP module, the ego-guided query is generated by cross-attending the ego status embedding with dense BEV features. These two queries are fused in the Fusion module based on the ego status embedding. The weight adapter also takes the ego status embedding as input. The encoded ego status is used throughout the three modules, with the encoding process shown in the center.}
\label{fig:model_structure}
\end{figure*}
\subsection{Autonomous Driving Planning}
Classical planning systems such as Apollo and Autoware rely on rule-based methods, including finite state machines (FSMs) and sampling-based search. These systems are interpretable and safe but struggle in complex real-world scenarios. Learning-based methods such as ChauffeurNet~\cite{bansal2019chauffeurnet} and Learning by Cheating (LBC)~\cite{Chen2019LBC} emerged to improve generalization by leveraging imitation or reinforcement learning, although they still depend on explicit perception and prediction inputs. 

The end-to-end paradigm started with models like PilotNet~\cite{PilotNet} and Conditional Imitation Learning (CIL)~\cite{CIL}, which directly predict control signals from camera images. However, these models suffer from instability due to the lack of intermediate semantic supervision. Later models, such as VectorNet~\cite{gao2020vectornet} incorporated map and actor interactions to improve trajectory quality. InterFuser~\cite{liu2022interfuser} enhanced perception robustness through sensor fusion, while ST-P3~\cite{hu2022stp3} combined map segmentation, occupancy prediction, and trajectory decoding into a multi-task system.

More recently, UniAD~\cite{hu2023uniad} proposed a unified query design to integrate multiple tasks into a single planning-centric model, achieving strong performance in perception, prediction, and planning. VAD~\cite{gu2023vad} introduces an end-to-end vectorized paradigm that simplifies the pipeline by leveraging fully vectorized scene representations, achieving state-of-the-art planning performance and improved inference speed. SparseDrive~\cite{sun2024sparsedrive} leverages the similarity between motion prediction and planning to design a parallel planning module, using sparse scene representation and collision-aware hierarchical trajectory selection. BridgeAD~\cite{zhang2025bridgead} bridged historical and future insights to achieve temporally consistent trajectories. However, these models either underutilize the online mapping module or fail to explicitly examine its role in end-to-end planning. In this work, we revisit the role of the online mapping module and explore how its outputs can be effectively assist current ego state information to enhance the robustness and performance of planning.

\section{Methodology}

\subsection{Overview}

The overall architecture of MAP is illustrated in Fig.~\ref{fig:model_structure}. It consists of four main components: the BEV (Bird’s Eye View) dense feature extraction module, the Plan-enhancing Online Mapping module, the Ego-status-guided Planning module, and the fusion module equipped with a learnable weight adapter. Together, these components form a cohesive system that integrates perception, semantic mapping, and trajectory planning in an end-to-end manner.

The process begins with BEVFormer, which takes as input multi-view images captured from onboard surround-view cameras and transforms them into a spatially aligned BEV representation. This representation serves as the visual backbone for downstream modules. In parallel, ego-vehicle status information, including velocity, acceleration, and heading angle obtained from the vehicle bus, as well as driving commands annotated in the dataset, is extracted to provide motion context. These features are subsequently fed into both the online mapping and trajectory planning modules, enabling them to generate more context-aware outputs.

Finally, the outputs from the mapping and planning pipelines are passed into the fusion module, where a learned weight adapter dynamically balances their contributions. The resulting fused query representation is decoded to produce the final trajectory, which reflects a comprehensive integration of perception, mapping, and current ego status information.

\subsection{Plan-enhancing Online Mapping Module}

In the baseline, the online mapping module outputs query representations that encode information about the drivable area, lane markings, crosswalks, and road boundaries. While such segmentation features may implicitly benefit planning, we enhance their utility by explicitly integrating them into the planning query. Specifically, we take the memory features $M_{\text{map}}$ produced by the online mapping module and apply cross-attention with the current ego vehicle status representation $E_{\text{ego}}$, generating a planning query $Q_{\text{map}}$ that directly contributes to the trajectory decoding stage.

\subsection{Ego-status-guided Planning Module}

Unlike the UniV2X baseline where all upstream queries are concatenated and fused via MLP before interacting with BEV features, we decouple trajectory planning into a dedicated module that operates in parallel with the online mapping pipeline. This module does not consume outputs from the online mapping module. Instead, it leverages both the BEV dense features $F_{\text{BEV}}$ and the ego-vehicle status $E_{\text{ego}}$.

The ego-vehicle status $E_{\text{ego}}$ is constructed from CAN-bus data, which provides the vehicle's global position and heading angle. Velocity and acceleration are computed from position differences across adjacent frames, normalized by the time intervals between frames. Driving command information is retrieved from a separate command dataset. The linear and nonlinear components of the ego-vehicle status are encoded separately and then concatenated to form the final input status vector.

Cross-attention between $F_{\text{BEV}}$ and the encoded ego status yields a planning query $Q_{\text{plan}}$, which is passed downstream for trajectory decoding.

\subsection{Fusion with Weight Adapter}

The fusion module receives two planning queries: $Q_{\text{map}}$ from the plan-enhancing online mapping module and $Q_{\text{plan}}$ from the ego-status-guided planning module. A learnable weight adapter is used to compute the fusion weight. It encodes the current ego vehicle status by applying an MLP to its linear components (e.g., velocity, acceleration, heading angle) and an embedding layer to its nonlinear part (e.g., command). These are concatenated and further encoded into a 64-dimensional feature vector, which is passed through a sigmoid activation to produce a weight coefficient $\alpha \in [0,1]$. The final fused query is computed as:
\[
Q_{\text{fused}} = \alpha \cdot Q_{\text{plan}} + (1 - \alpha) \cdot Q_{\text{map}}
\]
This fused representation is then decoded into a physical trajectory using a decoder, followed by a bivariate Gaussian activation to smooth the predicted trajectory points.

\subsection{Loss Functions}

The training objective consists of two primary components: online mapping loss and trajectory planning loss.

\paragraph{Online Mapping Loss.}  
The overall online mapping loss combines detection and segmentation supervision across multiple decoder layers. The final decoder layer supervises panoptic segmentation with separate losses computed for the things (foreground instances) and stuff (background semantics) categories, both using Dice loss for mask supervision. Auxiliary losses are computed for the first $L-1$ decoder layers to encourage multi-scale feature learning. Optionally, auxiliary losses from encoder outputs further improve detection performance. The model is trained with a sigmoid-based focal loss for classification, L1 loss for bounding box regression, and GIoU loss for bounding box localization. For segmentation, the mask loss is computed using Dice loss.
\begin{equation}
\mathcal{L}{\text{map}} = \mathcal{L}{\text{det}} + \mathcal{L}{\text{seg}} + \sum_{i=1}^{L-1} \mathcal{L}^{d_i}{\text{aux}}
\end{equation}
where
\begin{equation}
\mathcal{L}^{d_i}{\text{aux}} = \mathcal{L}^{d_i}{\text{det}} + \mathcal{L}^{d_i}{\text{seg}}
\end{equation}
Here, $\mathcal{L}{\text{det}}$ and $\mathcal{L}{\text{seg}}$ denote the detection and segmentation losses from the final decoder layer, respectively.

\begin{table*}[t]
\centering
\setlength{\tabcolsep}{3pt} 
\begin{tabular}{l|cccc|cccc|cccc|c}
\toprule
\multirow{2}{*}{\textbf{Method}} 
& \multicolumn{4}{c|}{L2 Error ($m$) ↓} 
& \multicolumn{4}{c|}{Col. Rate (\%) ↓} 
& \multicolumn{4}{c|}{Off-Road Rate (\%) ↓} 
& \multirow{2}{*}{Transm. Cos (BPS) ↓} \\
\cmidrule(lr){2-5} \cmidrule(lr){6-9} \cmidrule(lr){10-13}
 & 2.5s & 3.5s & 4.5s & Avg. 
 & 2.5s & 3.5s & 4.5s & Avg. 
 & 2.5s & 3.5s & 4.5s & Avg. 
 & \\
\midrule
CooperNaut~\cite{cui2022coopernaut} & 3.84 & 5.33 & 6.87 & 5.35 & 0.44 & 1.33 & 1.93 & 1.23 & \textbf{0.15} & \textbf{0.15} & 1.33 & \textbf{0.54} & $8.19 \times 10^7$ \\
UniV2X No Fusion & 2.58 & 3.37 & 4.36 & 3.44 & 0.15 & 1.04 & 1.48 & 1.08 & 0.44 & 0.56 & 2.22 & 1.08 & $0$ \\
UniV2X Vanilla & 2.33 & 3.69 & 5.12 & 3.71 & 0.59 & 2.07 & 3.70 & 2.12 & 0.15 & 1.33 & 4.74 & 2.07 & $8.19 \times 10^7$ \\
UniV2X BEV Fusion & 2.31 & 3.29 & 4.31 & 3.30 & \textbf{0.00} & 1.04 & 1.48 & 0.83 & 0.44 & 0.44 & 1.91 & 0.93 & $8.19 \times 10^7$ \\
UniV2X\textsuperscript{*}~\cite{univ2x} & 2.55 & 3.35 & 4.47 & 3.46 & \textbf{0.00} & \textbf{0.44} & \textbf{0.59} & \textbf{0.34} & 0.30 & 0.74 & 1.19 & 0.74 & $8.09 \times 10^5$ \\
UniV2X + Ego Status & 2.41 & 3.19 & 4.24 & 3.28 & 0.74 & 1.04 & 0.74 & 0.84 & 1.33 & 1.33 & 1.18 & 1.28 & $8.09 \times 10^5$ \\
\midrule 
\textbf{MAP (Ours)}  & \textbf{1.45} & \textbf{2.42} & \textbf{3.58} & \textbf{2.48} & \textbf{0.00} & 0.89 & 1.92 & 0.94 & \textbf{0.15} & 1.48 & \textbf{1.04} & 0.89 & $\mathbf{1.69 \times 10^5}$\textsuperscript{†} \\
\bottomrule
\end{tabular}
\caption{Comparison with baseline methods on the validation set under multi-step evaluation. * indicates that we re-trained the Stage Three: Cooperative Planning component on our own hardware. \textsuperscript{†}For MAP, we do not apply the same post-processing as in UniV2X, the transmission cost of occupancy probability maps is omitted.}
\label{tab:validation_results}
\end{table*}

\begin{table}[t]
\centering
\setlength{\tabcolsep}{3pt}
\small 
\begin{tabular}{lcccc}
\toprule
\textbf{Team} & \textbf{L2 (m)} $\downarrow$ & \textbf{Col. (\%)} $\downarrow$ & \textbf{Off. (\%)} $\downarrow$ & \textbf{Score} $\uparrow$ \\
\midrule
UniV2X & 3.07 & 0.71 & 0.89 & 0.591 \\
THU\_Song\textsuperscript{\dag} & 2.86 & 1.38 & 0.61 & 0.612 \\
\textbf{TJU\_Kan (Ours)*} & 2.67 & \textbf{0.67} & 0.46 & 0.841 \\
\textbf{TJU\_Kan (Ours)} & \textbf{2.56} & 0.96 & \textbf{0.39} & \textbf{0.854} \\
\bottomrule
\end{tabular}
\caption{
Comparison of models on the official leaderboard. \textsuperscript{\dag}~Second place on the leaderboard.
\textbf{TJU\_Kan (Ours)*} computes the ego status using a fixed time interval of 0.5s between frames, while \textbf{TJU\_Kan (Ours)} uses the actual time interval of each frame for more accurate computation. 
}
\label{tab:competition_results}
\end{table}

\begin{table*}[t]
\centering
\small
\begin{tabular}{l|cccc|cccc|cccc}
\toprule
\multirow{2}{*}{\textbf{Ablation study (on baseline)}} 
& \multicolumn{4}{c|}{\textbf{L2 Error (m) ↓}} 
& \multicolumn{4}{c|}{\textbf{Col. Rate (\%) ↓}} 
& \multicolumn{4}{c}{\textbf{Off-Road Rate (\%) ↓}} \\
\cmidrule(lr){2-5} \cmidrule(lr){6-9} \cmidrule(lr){10-13}
& 2.5s & 3.5s & 4.5s & Avg. 
& 2.5s & 3.5s & 4.5s & Avg. 
& 2.5s & 3.5s & 4.5s & Avg. \\
\midrule
Full UniV2X*               & 2.55 & 3.35 & 4.47 & 3.46 & \textbf{0.00} & \textbf{0.44} & \textbf{0.59} & \textbf{0.34} & \textbf{0.30} & \textbf{0.74} & 1.19 & \textbf{0.74} \\
w/o Track Pipeline & \textbf{2.53} & \textbf{3.32} & 4.45 & \textbf{3.43} & 0.44 & 0.74 & 1.04 & 0.74 & 0.44 & 0.89 & 1.33 & 0.89 \\
w/o Seg Pipeline      & 2.56 & 3.35 & 4.48 & 3.46 & 0.15 & \textbf{0.44} & 0.74 & 0.44 & \textbf{0.30} & \textbf{0.74} & \textbf{1.18} & \textbf{0.74} \\
w/o Motion Pipeline  & \textbf{2.53} & 3.33 & \textbf{4.44} & \textbf{3.43} & 0.30 & 0.74 & 1.18 & 0.74 & 0.59 & 0.89 & 1.33 & 0.94 \\
\bottomrule
\end{tabular}
\caption{Ablation study on UniV2X on the validation set by masking pipeline outputs under multi-step evaluation.* indicates that we re-trained the Stage Three: Cooperative Planning component on our own hardware.}
\label{tab:ablation}
\end{table*}

\begin{table*}[t]
\centering

\small
\begin{tabular}{l|cccc|cccc|cccc}
\toprule
\multirow{2}{*}{\textbf{Ablation study (on MAP)}} 
& \multicolumn{4}{c|}{\textbf{L2 Error (m) ↓}} 
& \multicolumn{4}{c|}{\textbf{Col. Rate (\%) ↓}} 
& \multicolumn{4}{c}{\textbf{Off-Road Rate (\%) ↓}} \\
\cmidrule(lr){2-5} \cmidrule(lr){6-9} \cmidrule(lr){10-13}
& 2.5s & 3.5s & 4.5s & Avg. 
& 2.5s & 3.5s & 4.5s & Avg. 
& 2.5s & 3.5s & 4.5s & Avg. \\
\midrule
w/o Plan-enhancing Online Mapping & 2.00 & 3.18 & 4.49 & 3.22 & \textbf{0.00} & 1.35 & 2.70 & 1.35 & \textbf{0.00} & 1.35 & 2.03 & 1.13 \\
w/o Ego-status-guided Planning              & 2.47 & 3.45 & 4.63 & 3.52 & 0.68 & \textbf{0.68} & \textbf{0.00} & \textbf{0.45} & \textbf{0.00} & 1.35 & 2.70 & 1.35 \\
Full MAP & \textbf{1.85} & \textbf{2.80} & \textbf{4.00} & \textbf{2.88} & 1.35 & 3.38 & 2.70 & 2.48 & \textbf{0.00} & \textbf{0.68} & \textbf{0.68} & \textbf{0.45} \\
\bottomrule
\end{tabular}
\caption{
Ablation study on our model. \textbf{To reduce training time, both the training and validation sets were subsampled to one-fifth of their original size.}
}
\label{tab:ablation_on_our_model}
\end{table*}

\paragraph{Trajectory Planning Loss.}
We build upon the planning loss from UniV2X and moved them from the Ego-status-guided Planning Module to the final supervision of the fused decoded trajectory $Q_{\text{fused}}$:

\begin{itemize}
\item \textbf{Collision Loss:}
For each future timestep $i$, the collision loss is computed by constructing the ego vehicle bounding box $\hat{\mathbf{B}}^{(i)}$ from the predicted position $(\hat{x}^{(i)}, \hat{y}^{(i)})$ extracted from $\hat{\mathbf{T}}^{(i)}$ and the ground truth heading $\theta^{(i)}{\text{gt}}$ taken from $\mathbf{T}{\text{gt}}^{(i)}$. The loss sums the overlapping areas between this bounding box and all obstacle bounding boxes at timestep $i$:

\begin{equation}
\ell_{\text{collision}}(\hat{\mathbf{T}}^{(i)}, \theta^{(i)}_{\text{gt}}) = \sum_{j=1}^{K_i} \Omega\left(\hat{\mathcal{B}}^{(i)}, \mathcal{B}_{j}^{(i)}\right),
\end{equation}

where $K_i$ is the number of obstacles at timestep $i$, and $j$ indexes these obstacles. The operator $\Omega(\cdot, \cdot)$ computes the intersection area between two bird’s-eye view bounding boxes by calculating the polygon formed by their overlapping corners. The bird’s-eye view bounding boxes are defined as:

\begin{equation}
\hat{\mathcal{B}}^{(i)} = \mathcal{B}\left(\hat{x}^{(i)}, \hat{y}^{(i)}, \theta^{(i)}_{\text{gt}}\right).
\end{equation}

The overall collision loss is

\begin{equation}
    \mathcal{L}_{\text{collision}} = \sum_{i=1}^{N} m^{(i)} \cdot \ell_{\text{collision}}(\hat{\mathbf{T}}^{(i)}, \theta^{(i)}_{\text{gt}}),
\end{equation}

where $m^{(i)}$ is a validity mask.

\item \textbf{Average Displacement Error(ADE) Loss:}  
The Average Displacement Error measures average Euclidean distance between predicted trajectory $\hat{\mathbf{T}}$ and ground truth $\mathbf{T}_{\text{gt}}$:  
\begin{equation}
\mathcal{L}_{\text{ADE}} = \frac{1}{ \sum_{t=1}^T m_t}  \sum_{t=1}^{T} m_t \left\|\hat{\mathbf{T}}_t - \mathbf{T}_{\text{gt},t}\right\|_2
\end{equation}

where $m_t$ is the validity mask for timestep $t$.

\item \textbf{Adaptive Loss:}  
To enhance the overall planning quality, we introduce an adaptive auxiliary loss that encourages the model to produce trajectories with lower displacement error, fewer collisions, and reduced off-road rate.:  
\begin{equation}
\mathcal{L}_{\text{adaptive}} = 
\alpha \cdot \mathcal{L}_{\text{L2}} + 
\beta \cdot \mathcal{L}_{\text{col}} + 
\gamma \cdot \mathcal{L}_{\text{off}}
\end{equation}

where $\alpha$, $\beta$, and $\gamma$ are manually set weight coefficients, and the calculation functions of $\mathcal{L}_{\text{L2}}$, $\mathcal{L}_{\text{col}}$, and $\mathcal{L}_{\text{off}}$ are obtained from validation statistics:
\begin{equation}
\mathcal{L}_{\text{L2}} = \sum_{t=1}^{T} m_t \cdot \left\| \hat{\mathbf{T}}_t - \mathbf{T}_{\text{gt},t} \right\|_2
\label{eq:l2}
\end{equation}

\begin{equation}
\mathcal{L}_{\text{col}} = \sum_{t=1}^{T} \mathbf{1} \left[ \hat{\mathbf{T}}_t \in \mathcal{O}_t \right] \cdot \mathbf{1} \left[ \mathbf{T}_{\text{gt},t} \notin \mathcal{O}_t \right]
\label{eq:collision}
\end{equation}

\begin{equation}
\mathcal{L}_{\text{off}} = \sum_{t=1}^{T} \mathbf{1} \left[ \hat{\mathbf{T}}_t \in \mathcal{U} \right]
\label{eq:out_of_drivable}
\end{equation}

where $\mathcal{O}_t$ denotes the obstacle region at time step $t$, $\mathcal{U}$ denotes the undrivable region, and $\mathbf{1}[\cdot]$ is the indicator function, equal to 1 if the condition holds and 0 otherwise.

\end{itemize}

\section{Experiments}

\subsection{Experimental Settings}

\textbf{Datasets and Metrics.} 
We evaluate our model on the DAIR-V2X-seq dataset, a large-scale, real-world V2X dataset designed for cooperative autonomous driving. This dataset contains 72,890 frames of synchronized 2D images and 3D point clouds with annotations, collected at 2Hz. The V2X-seq-SPD subset used in this challenge comprises over 15,000 sequential frames from 95 scenes and serves as the official benchmark for this task. 

According to the official evaluation protocol of the competition, three performance measures are used: L2 displacement error, collision rate, and off-road rate. For fair comparison, the leaderboard score is calculated as a normalized combination of these measures. The reference values $x_{\text{ref}}$ are: L2 error = 3.5 m, collision rate = 2\%, and off-road rate = 2.5\%. The corresponding improvement ranges $x_{\text{range}}$ are: L2 error = 1.0 m, collision rate = 1.5\%, and off-road rate = 2.5\%. Each metric score is calculated as:

\begin{equation}
\mathrm{score} = \frac{x_{\text{ref}} - x}{x_{\text{range}}}
\end{equation}

The overall score for this leaderboard will be the weighted average of these three normalized metrics, with weights of 0.5, 0.25, and 0.25, respectively.

\textbf{Training Details.} 
Our model predicts the future trajectory of ego vehicle for 5 seconds (10 timesteps). All models are trained using 2 NVIDIA RTX A800 GPUs. We initialize our model with the pretrained \texttt{univ2x\_coop\_e2e\_stg1} checkpoint from UniV2X and keep all hyperparameters consistent with UniV2X. The training process takes approximately 50 hours. 

\subsection{Comparison with Other Methods}
First, we evaluate our method against other models on the validation set. The results demonstrate that MAP effectively leverages the underutilized potential of online mapping in end-to-end systems. Compared to the UniV2X baseline, our method achieves a 28.3\% reduction in L2 displacement error (2.48~m vs. 3.46~m) on the DAIR-V2X-seq-SPD dataset (Table~\ref{tab:validation_results}), even without post-processing to constrain the trajectory within drivable areas. Meanwhile, it maintains a comparable off-road rate, which typically increases when L2 error is reduced. In addition, our architecture is simpler and more efficient, avoiding the need for heavy tracking, motion prediction, or occupancy grid modules, and thus significantly reducing memory usage, training time, and convergence steps. 

Furthermore, as shown in Table~\ref{tab:competition_results}, our model outperforms the baseline model on the leaderboard of the test set. Specifically, it achieves a 16.6\% reduction in L2 displacement error, a 56.2\% reduction in off-road rate, and a 44.5\% improvement in overall score. Notably, our model relies only on an online mapping module and an ego-status-guided planning module before fusion for trajectory decoding, thereby avoiding the complex multi-module dependencies of the baseline model. Moreover, unlike models such as AD-MLP~\cite{AD-MLP}, we do not rely on historical ego vehicle trajectories, which eliminates the dependence on past information and better aligns with the practical requirements of real-world autonomous driving.

\subsection{Ablation Studies}

Our ablation studies consist of two parts: one focusing on the baseline model and the other on our model.

\paragraph{Ablation Studies on baseline model.} Motivated by the complexity and resource consumption observed in the baseline model, we conduct a series of ablation studies to examine the contribution of each module. Surprisingly, as shown in Table~\ref{tab:ablation}, removing certain modules has minimal impact, prompting us to investigate similar findings in recent literature. Works such as~\cite{Li2024ego,paradrive} question the effectiveness of heavily stacked perception modules, and~\cite{Li2024ego} even suggests that the current ego status alone can yield competitive performance. These insights align with our observations: disabling the outputs of these modules leads to only marginal performance degradation. We suspect that this may be due to limitations in the original processing architecture, which fails to effectively utilize the outputs of individual modules. As a result, the generated planning trajectories inadequately integrate or misinterpret upstream information.

\paragraph{Ablation Studies on our model.} To further validate the effectiveness of our design, we conduct ablation studies on MAP. Given the simplicity of the architecture, we investigate the impact of removing (1) the outputs of the Plan-enhancing Online Mapping module and (2) the outputs of the Ego-status-guided Planning module. The results are reported in Table~\ref{tab:ablation_on_our_model}. As shown, each component contributes substantially to the final performance, highlighting their necessity. However, while the full MAP model demonstrates strong performance in L2 Error and Off-Road Rate, it shows relatively worse performance in Collision Rate. We hypothesize that this is mainly due to the removal of several perception modules from the original baseline, including the tracking module, motion prediction module, and occupancy prediction module. These components are omitted in our design to reduce memory consumption and model complexity, but their absence likely weakens the model’s ability to perceive surrounding agents accurately, leading to a higher collision rate. As future work, we plan to reintroduce selected modules to improve collision avoidance.

\section{Conclusion}

In this work, we present the Map-Assisted Planning (MAP) framework, a novel extension of the UniV2X baseline that fully leverages map resources for trajectory planning in V2X-enabled autonomous driving. By revisiting the often underutilized online mapping module, MAP explicitly integrates the current ego status with trajectory guidance extracted from the segmentation map. This design enables a dynamic fusion of spatial context and real-time ego status, producing planning outputs that are both accurate and robust. The proposed approach addresses the limitations of prior frameworks, where key intermediate modules were not fully exploited, and demonstrates the potential of map-centric information in improving end-to-end planning performance.

The official leaderboard shows that MAP achieves a \textbf{16.6\%} reduction in L2 displacement error, a \textbf{56.2\%} reduction in off-road rate, and a \textbf{44.5\%} improvement in overall score compared with the strong UniV2X baseline, all without any post-processing. Furthermore, MAP secures the \textbf{top ranking} in Track~2 of the End-to-End Autonomous Driving through V2X Cooperation Challenge at the MEIS Workshop @ CVPR~2025, surpassing the second-best model by \textbf{39.5\%} in overall score. These results highlight the critical value of map-based information for cooperative planning and point toward promising directions for designing the next generation of end-to-end autonomous driving systems.

\section*{Acknowledgement}

The authors would like to thank TÜV SÜD for their kind and generous support. We are also grateful to our colleagues at the Sino-German Center of Intelligent Systems for their valuable contributions. This work was additionally supported by the State Key Laboratory of Autonomous Intelligent Unmanned Systems and Frontiers Science Center for Intelligent Autonomous Systems, Ministry of Education.

{
    \small
    \bibliographystyle{ieeenat_fullname}
    \bibliography{main}

\begin{thebibliography}{24}
\providecommand{\natexlab}[1]{#1}
\providecommand{\url}[1]{\texttt{#1}}
\expandafter\ifx\csname urlstyle\endcsname\relax
  \providecommand{\doi}[1]{doi: #1}\else
  \providecommand{\doi}{doi: \begingroup \urlstyle{rm}\Url}\fi

\bibitem[Bansal et~al.(2019)Bansal, Krizhevsky, and Ogale]{bansal2019chauffeurnet}
Mayank Bansal, Alex Krizhevsky, and Abhijit Ogale.
\newblock Chauffeurnet: Learning to drive by imitating the best and synthesizing the worst.
\newblock In \emph{Proceedings of Robotics: Science and Systems (RSS)}, 2019.

\bibitem[Bojarski et~al.(2016)Bojarski, Testa, Dworakowski, Firner, Flepp, Goyal, Jackel, Monfort, Muller, Zhang, Zhang, Zhao, and Zieba]{PilotNet}
Mariusz Bojarski, Davide~Del Testa, Daniel Dworakowski, Bernhard Firner, Beat Flepp, Prasoon Goyal, Lawrence~D. Jackel, Mathew Monfort, Urs Muller, Jiakai Zhang, Xin Zhang, Jake Zhao, and Karol Zieba.
\newblock End to end learning for self-driving cars.
\newblock In \emph{arXiv preprint arXiv:1604.07316}, 2016.

\bibitem[Chen et~al.(2020)Chen, Zhou, Koltun, and Kr{\"a}henb{\"u}hl]{Chen2019LBC}
Dian Chen, Brady Zhou, Vladlen Koltun, and Philipp Kr{\"a}henb{\"u}hl.
\newblock Learning by cheating.
\newblock In \emph{Proceedings of the Conference on Robot Learning (CoRL)}, pages 66--75, 2020.
\newblock Spotlight Presentation.

\bibitem[Chen et~al.(2017)Chen, Zhu, Papandreou, Schroff, and Adam]{deeplabv2v3}
Liang-Chieh Chen, Yukun Zhu, George Papandreou, Florian Schroff, and Hartwig Adam.
\newblock Rethinking atrous convolution for semantic image segmentation.
\newblock \emph{arXiv preprint arXiv:1706.05587}, 2017.

\bibitem[Chen et~al.(2018{\natexlab{a}})Chen, Papandreou, Kokkinos, Murphy, and Yuille]{deeplabv1}
Liang-Chieh Chen, George Papandreou, Iasonas Kokkinos, Kevin Murphy, and Alan~L. Yuille.
\newblock Deeplab: Semantic image segmentation with deep convolutional nets, atrous convolution, and fully connected crfs.
\newblock \emph{IEEE Transactions on Pattern Analysis and Machine Intelligence}, 40\penalty0 (4):\penalty0 834--848, 2018{\natexlab{a}}.

\bibitem[Chen et~al.(2018{\natexlab{b}})Chen, Zhu, Papandreou, Schroff, and Adam]{deeplabv3plus}
Liang-Chieh Chen, Yukun Zhu, George Papandreou, Florian Schroff, and Hartwig Adam.
\newblock Encoder-decoder with atrous separable convolution for semantic image segmentation.
\newblock In \emph{Proceedings of the European Conference on Computer Vision (ECCV)}, pages 833--851, 2018{\natexlab{b}}.

\bibitem[Codevilla et~al.(2018)Codevilla, Müller, López, Koltun, and Dosovitskiy]{CIL}
Felipe Codevilla, Matthias Müller, Antonio López, Vladlen Koltun, and Alexey Dosovitskiy.
\newblock End-to-end driving via conditional imitation learning.
\newblock In \emph{Proceedings of the IEEE International Conference on Robotics and Automation (ICRA)}, pages 1--9, 2018.

\bibitem[Cui et~al.(2022)Cui, Qiu, Chen, Stone, and Zhu]{cui2022coopernaut}
Junning Cui, Han Qiu, Dian Chen, Peter Stone, and Yuke Zhu.
\newblock Coopernaut: End-to-end driving with cooperative perception for networked vehicles.
\newblock In \emph{Proceedings of the IEEE/CVF Conference on Computer Vision and Pattern Recognition (CVPR)}, pages 17252--17262, 2022.

\bibitem[Gao et~al.(2020)Gao, Wang, Wang, and Zhao]{gao2020vectornet}
Jiyang Gao, Yilun Wang, Yuxuan Wang, and Hang Zhao.
\newblock Vectornet: Encoding hd maps and agent dynamics from vectorized representation.
\newblock In \emph{Proceedings of the IEEE/CVF Conference on Computer Vision and Pattern Recognition (CVPR)}, pages 44--53, 2020.

\bibitem[Gu et~al.(2023)Gu, Wei, and Zhao]{gu2023vad}
Guangyu Gu, Zhiqiang Wei, and Hang Zhao.
\newblock Vad: Vectorized scene representation for efficient autonomous driving.
\newblock In \emph{Proceedings of the IEEE/CVF International Conference on Computer Vision (ICCV)}, pages 1--10, 2023.

\bibitem[Hu et~al.(2022)Hu, Chen, Wu, Li, Yan, and Tao]{hu2022stp3}
Shengchao Hu, Li Chen, Penghao Wu, Hongyang Li, Junchi Yan, and Dacheng Tao.
\newblock St-p3: End-to-end vision-based autonomous driving via spatial-temporal feature learning.
\newblock In \emph{Computer Vision – ECCV 2022}, pages 531--548. Springer, 2022.

\bibitem[Hu et~al.(2023)Hu, Yang, Chen, Wei, Zhang, Ma, Liu, Jiang, Chen, and Li]{hu2023uniad}
Yihan Hu, Jiazhi Yang, Li Chen, Zhiqiang Wei, Zheng Zhang, Xinzhu Ma, Yuwen Liu, Zhenghao Jiang, Qifeng Chen, and Hongyang Li.
\newblock Planning-oriented autonomous driving.
\newblock In \emph{Proceedings of the IEEE/CVF Conference on Computer Vision and Pattern Recognition (CVPR)}, pages 17853--17862, 2023.

\bibitem[Li et~al.(2022{\natexlab{a}})Li, Wang, Wang, and Zhao]{HDMapNet}
Qi Li, Yilun Wang, Yuxuan Wang, and Hang Zhao.
\newblock {HDMapNet}: An online {HD} map construction and evaluation framework.
\newblock In \emph{Proceedings of the IEEE International Conference on Robotics and Automation (ICRA)}, pages 4628--4634, 2022{\natexlab{a}}.

\bibitem[Li et~al.(2022{\natexlab{b}})Li, Wang, Xie, Yu, Anandkumar, Alvarez, Luo, and Lu]{panopticsegformer2022}
Zhiqi Li, Wenhai Wang, Enze Xie, Zhiding Yu, Anima Anandkumar, Jose~M. Alvarez, Ping Luo, and Tong Lu.
\newblock Panoptic segformer: Delving deeper into panoptic segmentation with transformers.
\newblock In \emph{Proceedings of the IEEE/CVF Conference on Computer Vision and Pattern Recognition (CVPR)}, pages 1280--1289, 2022{\natexlab{b}}.

\bibitem[Li et~al.(2024)Li, Yu, Lan, Li, Kautz, Lu, and Alvarez]{Li2024ego}
Zhiqi Li, Zhiding Yu, Shiyi Lan, Jiahan Li, Jan Kautz, Tong Lu, and Jose~M. Alvarez.
\newblock Is ego status all you need for open-loop end-to-end autonomous driving?
\newblock In \emph{Proceedings of the IEEE/CVF Conference on Computer Vision and Pattern Recognition (CVPR)}, pages 14864--14873, 2024.

\bibitem[Liao et~al.(2023)Liao, Chen, Zhang, Jiang, Cheng, Zhang, Liu, Huang, and Wang]{Liao2023MapTR}
Bencheng Liao, Shaoyu Chen, Yunchi Zhang, Bo Jiang, Tianheng Cheng, Qian Zhang, Wenyu Liu, Chang Huang, and Xinggang Wang.
\newblock Maptr: End-to-end transformer for vectorized hd map construction.
\newblock In \emph{Proceedings of the International Conference on Learning Representations (ICLR)}, 2023.

\bibitem[Liu et~al.(2023)Liu, Yuan, Wang, Wang, and Zhao]{liu2022vectormapnet}
Yicheng Liu, Tianyuan Yuan, Yue Wang, Yilun Wang, and Hang Zhao.
\newblock Vectormapnet: End-to-end vectorized hd map learning.
\newblock In \emph{International conference on machine learning (ICML)}, 2023.

\bibitem[Shao et~al.(2022)Shao, Wang, Chen, Li, and Liu]{liu2022interfuser}
Hao Shao, Letian Wang, RuoBing Chen, Hongsheng Li, and Yu Liu.
\newblock Safety-enhanced autonomous driving using interpretable sensor fusion transformer.
\newblock In \emph{Proceedings of the Conference on Robot Learning (CoRL)}, pages 1--10, 2022.

\bibitem[Sun et~al.(2025)Sun, Lin, Shi, Zhang, Wu, and Zheng]{sun2024sparsedrive}
Wenchao Sun, Xuewu Lin, Yining Shi, Chuang Zhang, Haoran Wu, and Sifa Zheng.
\newblock Sparsedrive: End-to-end autonomous driving via sparse scene representation.
\newblock In \emph{Proceedings of the IEEE International Conference on Robotics and Automation (ICRA)}, 2025.

\bibitem[Weng et~al.(2024)Weng, Ivanovic, Wang, Wang, and Pavone]{paradrive}
Xinshuo Weng, Boris Ivanovic, Yan Wang, Yue Wang, and Marco Pavone.
\newblock Para-drive: Parallelized architecture for real-time autonomous driving.
\newblock In \emph{Proceedings of the IEEE/CVF Conference on Computer Vision and Pattern Recognition (CVPR)}, pages 15449--15458, 2024.

\bibitem[Yu et~al.(2024)Yu, Yang, Zhong, Yang, Fan, Luo, and Nie]{univ2x}
Haibao Yu, Wenxian Yang, Jiaru Zhong, Zhenwei Yang, Siqi Fan, Ping Luo, and Zaiqing Nie.
\newblock End-to-end autonomous driving through v2x cooperation.
\newblock In \emph{Proceedings of the AAAI Conference on Artificial Intelligence}, pages 9598--9606, 2024.

\bibitem[Zhai et~al.(2023)Zhai, Feng, Du, Mao, Liu, Tan, Zhang, Ye, and Wang]{AD-MLP}
Jiang-Tian Zhai, Ze Feng, Jinhao Du, Yongqiang Mao, Jiang-Jiang Liu, Zichang Tan, Yifu Zhang, Xiaoqing Ye, and Jingdong Wang.
\newblock Rethinking the open-loop evaluation of end-to-end autonomous driving in nuscenes.
\newblock \emph{arXiv preprint arXiv:2305.10430}, 2023.

\bibitem[Zhang et~al.(2025)Zhang, Song, Jin, and Zhang]{zhang2025bridgead}
Bozhou Zhang, Nan Song, Xin Jin, and Li Zhang.
\newblock Bridging past and future: End-to-end autonomous driving with historical prediction and planning.
\newblock In \emph{Proceedings of the IEEE/CVF Conference on Computer Vision and Pattern Recognition (CVPR)}, 2025.

\bibitem[Zhao et~al.(2017)Zhao, Shi, Qi, Wang, and Jia]{PSPNet}
Hengshuang Zhao, Jianping Shi, Xiaojuan Qi, Xiaogang Wang, and Jiaya Jia.
\newblock Pyramid scene parsing network.
\newblock In \emph{Proceedings of the IEEE Conference on Computer Vision and Pattern Recognition (CVPR)}, pages 2881--2890, 2017.

\end{thebibliography}
}

\end{document}